\documentclass[lettersize,journal]{IEEEtran}

\usepackage{amssymb,amsmath}
\usepackage{cite}
\usepackage{graphicx}
\usepackage{subfig}

\usepackage{psfrag}
\usepackage{url}
\usepackage[latin1]{inputenc}
\usepackage[absolute,overlay]{textpos}
\usepackage{gensymb}
\usepackage{cases}
\usepackage[font=footnotesize]{caption}
\usepackage{float}
\usepackage[linesnumbered,ruled,lined]{algorithm2e}
\usepackage{pgf}

\usepackage[nocomma]{optidef}

\usepackage{amsthm}

\usepackage{booktabs}
\usepackage{color,soul}

\usepackage{textcomp}
\usepackage{bbm}
\usepackage{xcolor}
\def\BibTeX{{\rm B\kern-.05em{\sc i\kern-.025em b}\kern-.08em
    T\kern-.1667em\lower.7ex\hbox{E}\kern-.125emX}}

\usepackage{tabularx}
\usepackage{makecell}
\usepackage{multirow}

\DeclarePairedDelimiter\ceil{\lceil}{\rceil}

\begin{document}

\title{Semantic Meta-Split Learning: A TinyML Scheme for Few-Shot Wireless Image Classification}
\author{
	\IEEEauthorblockN{Eslam Eldeeb, Mohammad Shehab, Hirley Alves, and Mohamed-Slim Alouini \\
	}
	\thanks{Eslam Eldeeb and Hirley Alves are with Centre for Wireless Communications (CWC), University of Oulu, Finland. 
                (e-mail: firstname.lastname@oulu.fi). 
                Mohammad Shehab and Mohamed-Slim Alouini are with CEMSE Division, King Abdullah University of Science and Technology (KAUST), Thuwal 23955-6900, Saudi Arabia (email: mohammad.shehab@kaust.edu.sa, slim.alouini@kaust.edu.sa).
        }
	\thanks{The work of E. Eldeeb and H. Alves was partially supported by the Research Council of Finland (former Academy of Finland)
                6G Flagship Programme (Grant Number: 346208) and by the European Commission through the Hexa-X-II (GA no. 101095759). 
        }
}
\maketitle

\begin{abstract}
Semantic and goal-oriented (SGO) communication is an emerging technology that only transmits significant information for a given task. Semantic communication encounters many challenges, such as computational complexity at end users, availability of data, and privacy-preserving. This work presents a TinyML-based semantic communication framework for few-shot wireless image classification that integrates split-learning and meta-learning. We exploit split-learning to limit the computations performed by the end-users while ensuring privacy-preserving. In addition, meta-learning overcomes data availability concerns and speeds up training by utilizing similarly trained tasks. The proposed algorithm is tested using a data set of images of hand-written letters. In addition, we present an uncertainty analysis of the predictions using conformal prediction (CP) techniques. Simulation results show that the proposed Semantic-MSL outperforms conventional schemes by achieving $20 \%$ gain on classification accuracy using fewer data points, yet less training energy consumption. 
\end{abstract}

\section{Introduction}\label{sec:introduction}


The road to intelligent communication systems encompasses a paradigm shift from conventional communication models to SGO communication frameworks~\cite{zhang2022goal}. SGO communication involves the transmission of only the essential semantics of a message instead of the entire message, which helps to conserve power and improve the spectral efficiency of the communication system~\cite{qin2021semantic}. Semantic communication has shown great promise in a wide range of applications. For instance, it has been used for efficient wireless image transmission and classification~\cite{bourtsoulatze2019deep}, natural language processing~\cite{sheng2022multi}, speech processing~\cite{weng2021semantic}, and multi-modal fusion~\cite{xie2021task}.

Semantic communication enables the receiver to handle various tasks by using only the relevant data that is useful for each specific task. This approach results in more efficient data compression, less signaling overhead and energy consumption, and higher spectral efficiency for the network. However, current GO communication models faces many challenges related to data collection~\cite{9144301,eldeeb2024conservativeriskawareofflinemultiagent} privacy preserving~\cite{shen2021data}, and amount of computations at the end users~\cite{kairouz2021advances}. For instance, conventional deep neural network (DNN) models rely on large amounts of data for training, which might not be feasible in many applications due to the high cost of data collection, the scarcity of resources available for data transmission or privacy concerns~\cite{agarwal2020optimistic}. In that case, the model usually suffers from high uncertainties due to insufficient data required to build, train and validate the model.

Meanwhile, semantic communication approaches rely on heavy and power-hungry machine learning (ML) and DNNs models for feature extraction and handling large volumes of data. In addition, various use-cases have privacy restrictions on data transmission~\cite{liang2008secure}, whereas other applications have low-power / low-resources at end users that make it challenging to perform complex deep learning computations~\cite{singh2019detailed,9845353}. To this end, TinyML aims at inventing light, yet accurate and communication-efficient ML schemes that consume fewer energy resources to achieve the learning goals \cite{Tiny1,Tiny2}.


This paper presents an energy and communication efficient TinyML scheme for wireless image transmission classification, termed Semantic Meta-Split Learning or simply \emph{Semantic-MSL}. The proposed method integrates split-learning~\cite{gao2020end} and meta-learning~\cite{finn2017modelagnostic} to perform few-shot over-the-air (OTA) image classification while ensuring efficient data compression, low computation complexity, and energy consumption at the end users as well as privacy-preservation. To evaluate the performance of the proposed model, we test the algorithm using a data set of images of hand-written letters used for classification.

\begin{figure}[t!]
    \centering
    \includegraphics[width=1\columnwidth,trim={0 0 0 0},clip]{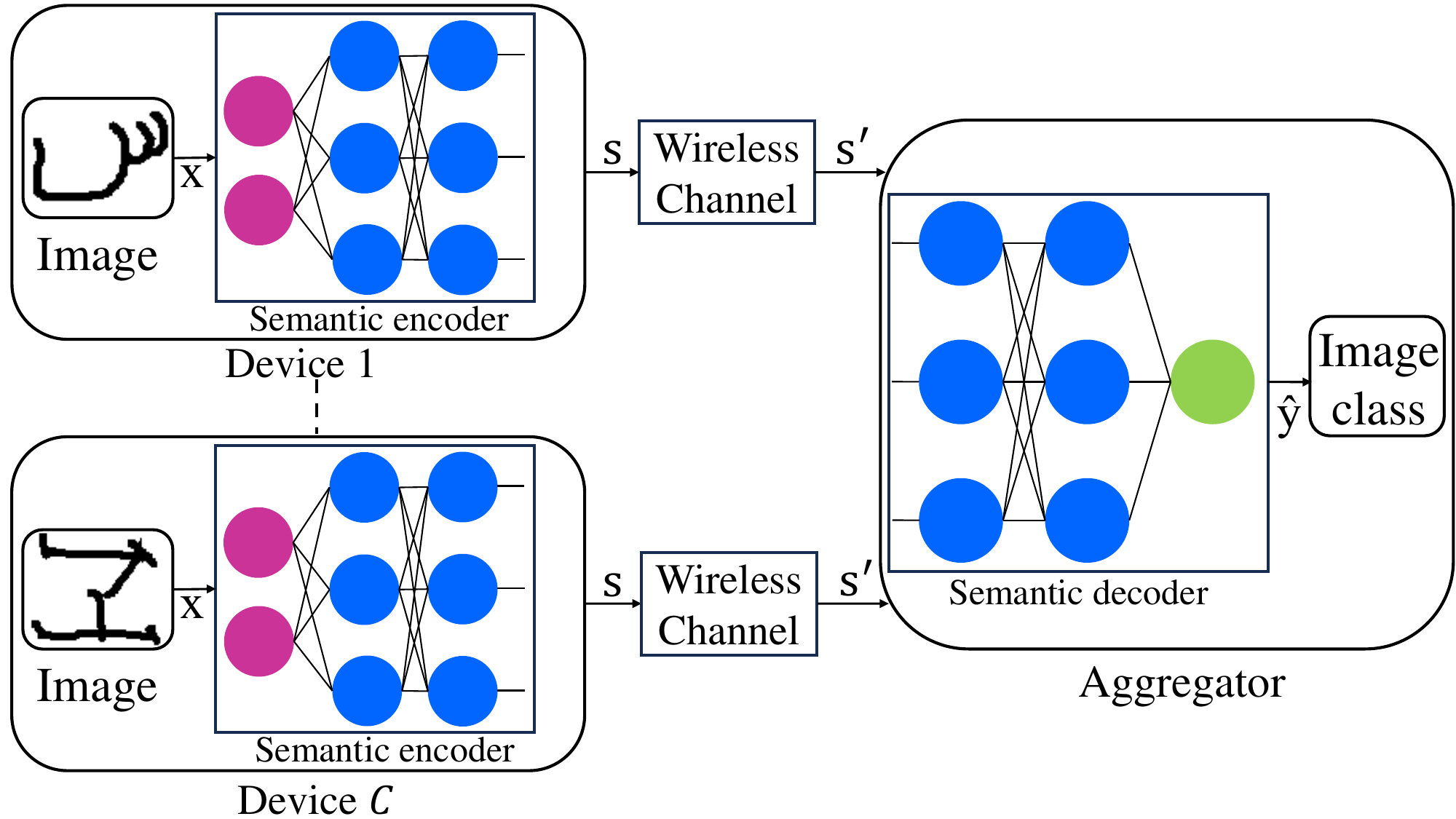} 
    \caption{Multiple devices aim to classify the correct letters in images sampled from different languages. Each device transmits the semantics of the image through a wireless channel to the aggregator, which predicts the image class and transmits it back to the device.} 
    \label{System_Model}
\end{figure}

\subsection{Split-Learning and Meta-Learning}

\emph{Split-learning}~\cite{gao2020end} is a family of collaborative and distributed ML algorithms, where the learning model, i.e., neural network, is divided into \emph{device-side model} and \emph{aggregator-side model} separated by the \emph{cut-layer} as shown in Fig.~\ref{System_Model}. The training and testing of the model are executed among the devices and the aggregator. During forward propagation, the devices transmit only the output of their last layer, i.e., \emph{smashed data}, to the aggregator, which, in turn, transmits back the \emph{smashed data's gradients} during back-propagation~\cite{da2024distributed}.

The major advantage of split learning over federated learning is that it reduces the heavy computations performed at the device side in federated learning and the high signaling overhead resulting from transmitting the whole model to the aggregator. In split learning, devices, and aggregators only exchange smashed data and gradients~\cite{10040976}, which saves communication energy as required by TinyML techniques. Moreover, split learning provides high levels of privacy-preserving as the data is not shared between the devices and the aggregator, which can access only a portion of the model.

In addition to split-learning, the proposed framework exploits \emph{meta-learning}~\cite{finn2017modelagnostic}, which is known as \emph{learning-to-learn}. Conventional deep learning requires the model's training from scratch for each new configuration adjusted in the network. In contrast, meta-learning is motivated by utilizing different configurations (tasks) to infer a model that performs well in new configurations. In this context, \emph{model-agnostic meta-learning (MAML)} is a famous family of meta-learning that meta-trains multiple tasks searching for suitable initialization for the weights of the trainable model so that the model converges in a few gradient steps using a few shots of data points. Meta-learning has shown promising performance in different wireless communication use cases, such as channel estimation, symbol demodulation, and modulation classification~\cite{issa2023metalearning,park2020meta}.

\subsection{Related Work}

The advances in semantic communication have been addressed extensively in many recent works. Specifically, the work in~\cite{raha2023artificial} proposes a framework that enables the transmission of detected traffic signs from one autonomous vehicle to another. The authors in~\cite{zhang2023wireless} propose a convolutional neural network (CNN) architecture for wireless image transfer while ensuring security awareness. They focus on privacy preservation without tackling the problem of data availability or energy consumpation. In~\cite{9830752}, Xie et al. propose transformers-based models for multiple task-oriented communication, whereas the work in~\cite{9763856} discusses resource allocation techniques for semantic communication. In contrast to our work, the proposed methods lack accumulating experience from various tasks (i.e, meta-learning) for training.

Recently, split-learning has also gained popularity in the field of wireless communication. For instance, in the article~\cite{lin2024efficient}, the authors have introduced a split-learning scheme where learning is carried out across various devices in a parallel manner. However, they have only optimized the cut layer position without considering the uncertainty of different model configurations. The authors of\cite{10038613} poropose a MIMO-based OTA split learning approach that interestingly also estimates the channel efficienty through forward and backward propagation processes improving the overall system efficiency. Meanwhile in~\cite{9923620}, a combination of split and federated learning is utilized for unmanned aerial vehicle (UAV) applications. In~\cite{10304624}, federated learning has been combined with split learning to reduce communication latency and improve accuracy within heterogenous devices.

Meta-learning has gained further attention in wireless communication applications. The work in~\cite{10293257} exploits meta-learning with conformal prediction (CP) to design a well-calibrated model via a few shots of training data. In~\cite{9758695}, Zhang et al. combine meta-learning and federated learning for wireless traffic prediction. The authors in~\cite{9457160} optimize the trajectories of multiple UAVs using meta-learning, whereas the authors in~\cite{9322414} optimize the UAV trajectory using meta-reinforcement learning. To our knowledge, this is the first work to combine split-learning and meta-learning for goal-oriented semantic communications and few-shot wireless image classification.

\subsection{Contribution}
This paper introduces a novel TinyML-based semantic communication framework that combines meta-learning and split-learning for efficient wireless image transmission and classification. The main contributions of the paper are summarized as follows:
\begin{itemize}
    \item We propose the Semantic-MSL scheme architecture using convolutional neural networks (CNNs). Split learning ensures low computations on the device side, and meta-learning ensures fast learning using a small amount of data. 

    \item To showcase the proposed framework, we assume a few-shot classification (i.e., only a few images are available for training) of hand-written letters being transmitted between devices and an aggregator.
 
    \item We exploit conformal prediction to quantify the uncertainty associated with the proposed model. Simulation results show that the proposed Semantic-MSL framework significantly outperforms both SL and conventional deep learning in terms of classification accuracy, conformal prediction uncertainty, and energy consumption.

    \item We show the effect of changing the position of the cut layer on the required computational power at the users, the deployment time of the algorithm, and the amount of information transmitted.
\end{itemize}

The rest of the paper is organized as follows: Section~\ref{sec:sysmodel} describes the system model and defines the problem. Section~\ref{Sec:Background} introduces split-learning and meta-learning in brief. In section~\ref{sec:proposed}, we present the proposed Semantic-MSL communication framework. Section~\ref{Sec:KPI} explains the performance evaluation metrics, including conformal prediction, whereas section~\ref{sec:results} shows the experimental results. Section~\ref{sec:conclusions} concludes the paper. The main symbols and acronyms used in this paper are summarized in TABLE \ref{tab:acronyms}.

\begin{table}[ht]
\centering
\caption{Important abbreviations and sybmols}
\label{tab:acronyms}
\begin{tabular}{cl}
\toprule
\textbf{Abbreviations} & \textbf{Definition} \\ 
\midrule \midrule
CNN          & Convolutional neural network \\ \hline
CP           & Conformal Predicition \\ \hline
MAML          & Model agnostic meta-leanring \\ \hline
MSL           & Meta-split learning \\ \hline
NC           & Nonconformity \\ \hline
OTA          & over the air \\ \hline
SGD           & Stochastic gradient descent\\ \hline
SGO            & Semantic and Goal Oriented\\ \hline
TinyML           & Tiny Machine Learning \\ \hline

\bottomrule
& \\
\textbf{Symbol} & \textbf{Description} \\ 
\midrule \midrule
$\mathcal{C} $             & Set of users\\ \hline 
$\texttt{cov}(\cdot) $             & Coverage function\\ \hline 
$M $             & number of images in each class\\ \hline 
$\mathbf{s} $             & Transmitted semantic message / smashed data\\ \hline
$ \mathbf{s^{\prime}} $       & Received semantic message\\ \hline 
$\mathcal{T} $             & Set of tasks\\ \hline 
$\mathbf{W^C} $             & Parameter vector of the device\\ \hline 
$\mathbf{W^S} $             & Parameter vector of the aggregator \\ \hline 
$\mathcal{Y} $             & Set of image classes \\ \hline 
$ \mathbf{y}$             & True image class\\ \hline 
$ \hat{\mathbf{y}}$             & Predicted image class\\ \hline

$\beta $             & Meta-learning rate\\ \hline 
$\eta $             & Semantic learning rate\\ \hline 
$\Gamma $             & Prediction set\\ \hline 
$\nabla $             & Gradient operator\\ \hline 

\bottomrule
\end{tabular}
\end{table}

\section{System Model and Problem Formulation}\label{sec:sysmodel}

\subsection{System model}

As shown in Fig.~\ref{System_Model}, consider a set of users $\mathcal{C} = \{1, 2, \cdots, \mathbf{C}\}$, where $\mathbf{C} = |\mathcal{C}|$. Each user transmits an image $\mathbf{x}$ to an aggregator (\textit{e.g.}, BS) through a wireless channel. Consider a task distribution $p(\tau)$ and $T$ tasks forming the set $\mathcal{T} = \{\tau_1, \cdots, \tau_T\}$, where each user $c$ samples a task from the set of tasks $\mathcal{T}$. Each task composes of multiple image classes forming the set $\mathcal{Y} = \{1, 2, \cdots, \mathbf{Y}\}$, where $\mathbf{Y} = |\mathcal{Y}|$. Each class contains $M = |\mathcal{M}|$ images defined as $\mathcal{X} = \{\mathbf{x}_1, \mathbf{x}_2, ..., \mathbf{x}_M\}$. On the user side, the semantic encoder encodes the image as
\begin{equation}
    \mathbf{s} = f_C(\mathbf{x}),
\end{equation}
where $\mathbf{s} \in \mathbb{R}^{N \times 1}$ is the transmitted semantic message, $N$ is the size of the message, and $f_C(\cdot)$ is the semantic encoder~\cite{eldeeb2024multi}. The encoded message is modulated using Quadrature amplitude modulation (QAM) modulation and transmitted to the receiver. The transmitted signal has a power constraint $\frac{1}{N} \mathbb{E} \{ \mathbf{x^2} \} \leq p$.

Consider an additive white Gaussian noise (AWGN) channel with noise distribution $\mathbf{n} \sim \mathcal{N} (0,\sigma^2 \mathbf{I})$, where $\sigma^2$ is the noise power
\begin{equation}
    \mathbf{s^{\prime}} = \mathbf{h} \: \mathbf{s} + \mathbf{n},
\end{equation}
where $\mathbf{h}$ is the channel component that comprises a Rayleigh flat fading. At the aggregator, the semantic message is decoded through the semantic decoder to depict the class of the image
\begin{equation}
    \hat{\mathbf{y}} = f_S(\mathbf{s^{\prime}}),
\end{equation}
where $f_S(\cdot)$ is the semantic decoder. 

\subsection{Problem Definition}
The main objective is to estimate the true class of each image using only the transmitted semantic information of the image. In addition, we aim to perform this estimation with the lowest possible number of images. In particular, the target is to find the best models $f_C(\cdot)$ and $f_S(\cdot)$ so that the true classes of the images can be estimated using a few shots of training images. This optimization problem is formulated as
\begin{subequations}\label{P1}
	\begin{alignat}{2}
	\mathbf{P1:}\: \: \: \: &\underset{\mathbf{s},K}{\min}   &\ &  \sum_{T} \bigg[ \sum_{K} \mathcal{L}(\mathbf{y},\hat{\mathbf{y}}) \bigg] + \zeta \: K \: \mathbf{Y}, 
	\ \\
	&\text{s.t.}   &      & K \leq M,\label{P1:a}\\
		& & & \hat{\mathbf{y}}, \mathbf{y} \in \mathcal{Y}, \label{P1:b}
	\end{alignat}
\end{subequations}
where $\mathcal{L}(\mathbf{y},\hat{\mathbf{y}})$ is a loss function that measures the distance between the true class and the predicted class, $K$ represents the number of images sampled from each task, and $\zeta$ is a weighting factor. The constraint \eqref{P1:a} forces the number of sampled images $K$ to be lower than or equal to the total number of images available in that class. In contrast, the constraint \eqref{P1:b} ensures that the output $\mathbf{y}$ and the estimated output $\hat{\mathbf{y}}$ are in the set of all possible classes.

\section{Background}\label{Sec:Background}
This section presents an overview of split learning and meta-learning. This discussion is necessary to prepare for the proposed meta-split learning-based semantic communication system for wireless transmission and few-shot classification of images.

\subsection{Split-Learning}

\begin{figure}[t!]
    \centering
    \includegraphics[width=1\columnwidth,trim={0 0 0 0},clip]{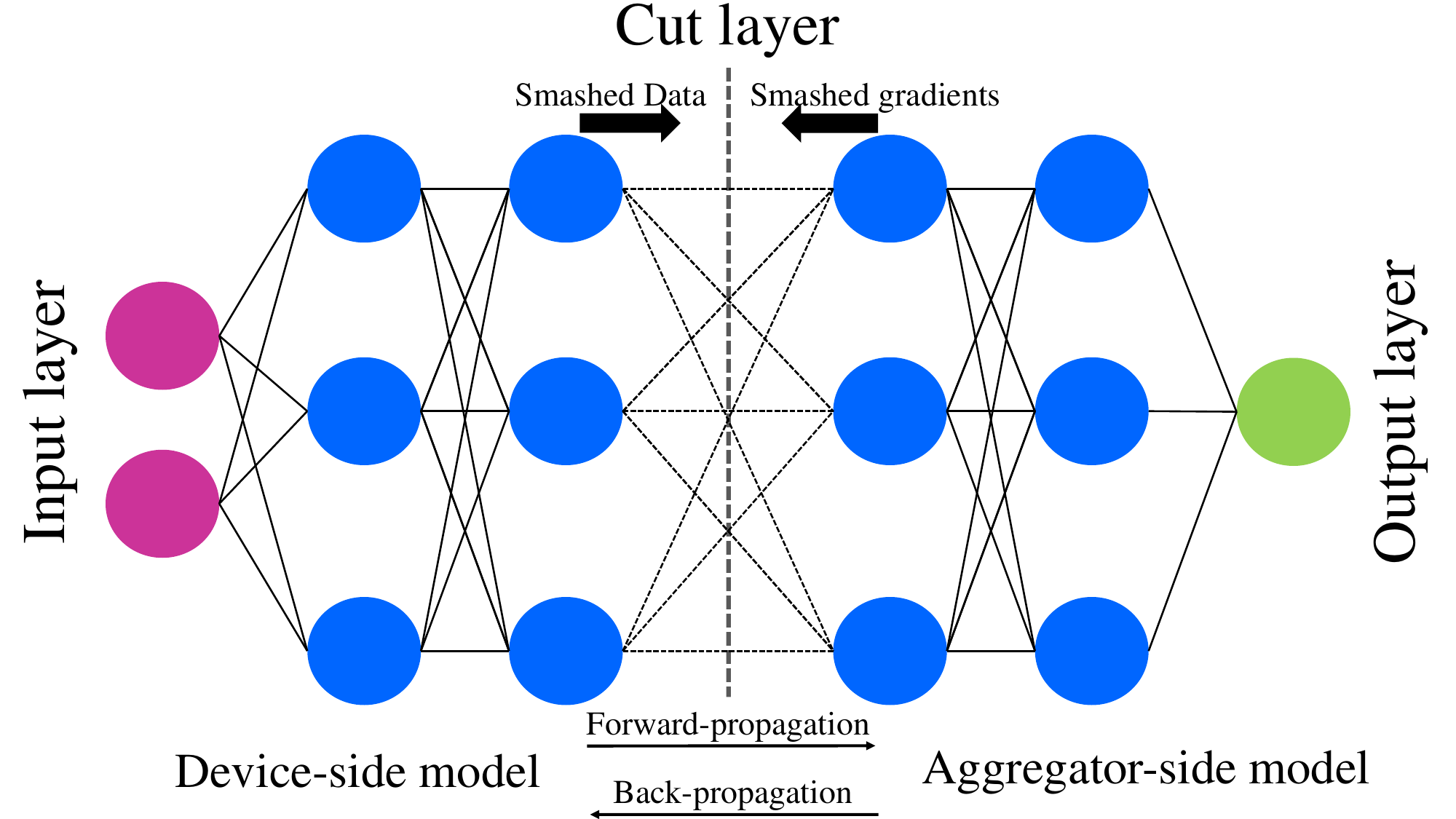} \vspace{2mm}
    \caption{Illustration of the architecture of split learning. The model is divided at the cut layer into a device-side model and an aggregator-side model, where the training is taking part at both entities.} 
    \vspace{0mm}
    \label{Split_Learning}
\end{figure}

\emph{Split-learning}~\cite{gao2020end} is a distributed machine learning framework that divides the machine learning architecture into multiple sectors, i.e., a neural network. These sectors are distributed among multiple devices and aggregators. As shown in Fig.~\ref{Split_Learning}, the sector at the device is referred to as \emph{device-side model} with $\mathbf{W^C}$ parameters vector, whereas the sector at the aggregator is called \emph{aggregator-side model} with $\mathbf{W^S}$ parameters vector. The layer dividing the device-side and aggregator-side models is called the \emph{cut layer}. Consider a model whose input $\mathbf{x}$ corresponds to a target output $\mathbf{y}$. Herein, the input layer exists at the device side, where the input is forwarded through the device-side model until the cut layer. The output of the device-side model is called \emph{smashed data $\mathbf{s}$}, which is transmitted to the aggregator-side model
\begin{equation}
\label{Client_model_eq}
    \mathbf{s} = \mathbf{W^C} \: \mathbf{x}.
\end{equation}

Ideally, the smashed data is received at the aggregator side and considered the input to the aggregator-side model. The smashed data propagates through the aggregator-side model, giving 
\begin{equation}
\label{Server_model_eq}
    \hat{\mathbf{y}} = \mathbf{W^S} \: \mathbf{s},
\end{equation}
where $\hat{\mathbf{y}}$ is the output of the aggregator-side model. Without loss of generality, the output $\hat{\mathbf{y}}$ is compared to the target output $\mathbf{y}$ using an error measure function, i.e., an arbitrary loss function $\mathcal{L}_{\mathbf{W^C},\mathbf{W^S}} (\mathbf{y},\hat{\mathbf{y}})$. The parameters of the aggregator-side model are updated using stochastic gradient descent (SGD)
\begin{equation}
\label{update_server_param}
    \mathbf{W^S} \leftarrow \mathbf{W^S} - \eta \nabla_{\mathbf{W^S}}\mathcal{L}_{\mathbf{W^C},\mathbf{W^S}} (\mathbf{y},\hat{\mathbf{y}}),
\end{equation}
where $\eta$ is the semantic learning rate and $\nabla_{\mathbf{W^S}}\mathcal{L}_{\mathbf{W^C},\mathbf{W^S}} (\mathbf{y},\hat{\mathbf{y}})$ is the gradient calculated concerning the aggregator-side parameters until the cut-layer.
Afterward, the smashed data's gradients are transmitted back to the device-side model, where the parameters of the device-side model are updated via
\begin{equation}
\label{update_client_param}
    \mathbf{W^C} \leftarrow \mathbf{W^C} - \eta \nabla_{\mathbf{W^C}}\mathcal{L}_{\mathbf{W^C},\mathbf{W^S}} (\mathbf{y},\hat{\mathbf{y}}),
\end{equation}
where $\nabla_{\mathbf{W^C}}\mathcal{L}_{\mathbf{W^C},\mathbf{W^S}} (\mathbf{y},\hat{\mathbf{y}})$ is the gradient calculated concerning the device-side parameters (i.e, the way back to the input). This procedure is repeated until convergence.

Split learning reduces the computational load at the devices by dividing the model between the devices and the aggregator~\cite{thapa2020advancements}. The reduction depends on the position of the cut layer. The earlier the cut layer, the lower the computational load on the device. However, this comes at the cost of a larger size of the transmitted message and lower information security. 


\subsection{Meta-Learning}

\begin{figure}[t!]
    \centering
    \includegraphics[width=1\columnwidth,trim={0 0 0 0},clip]{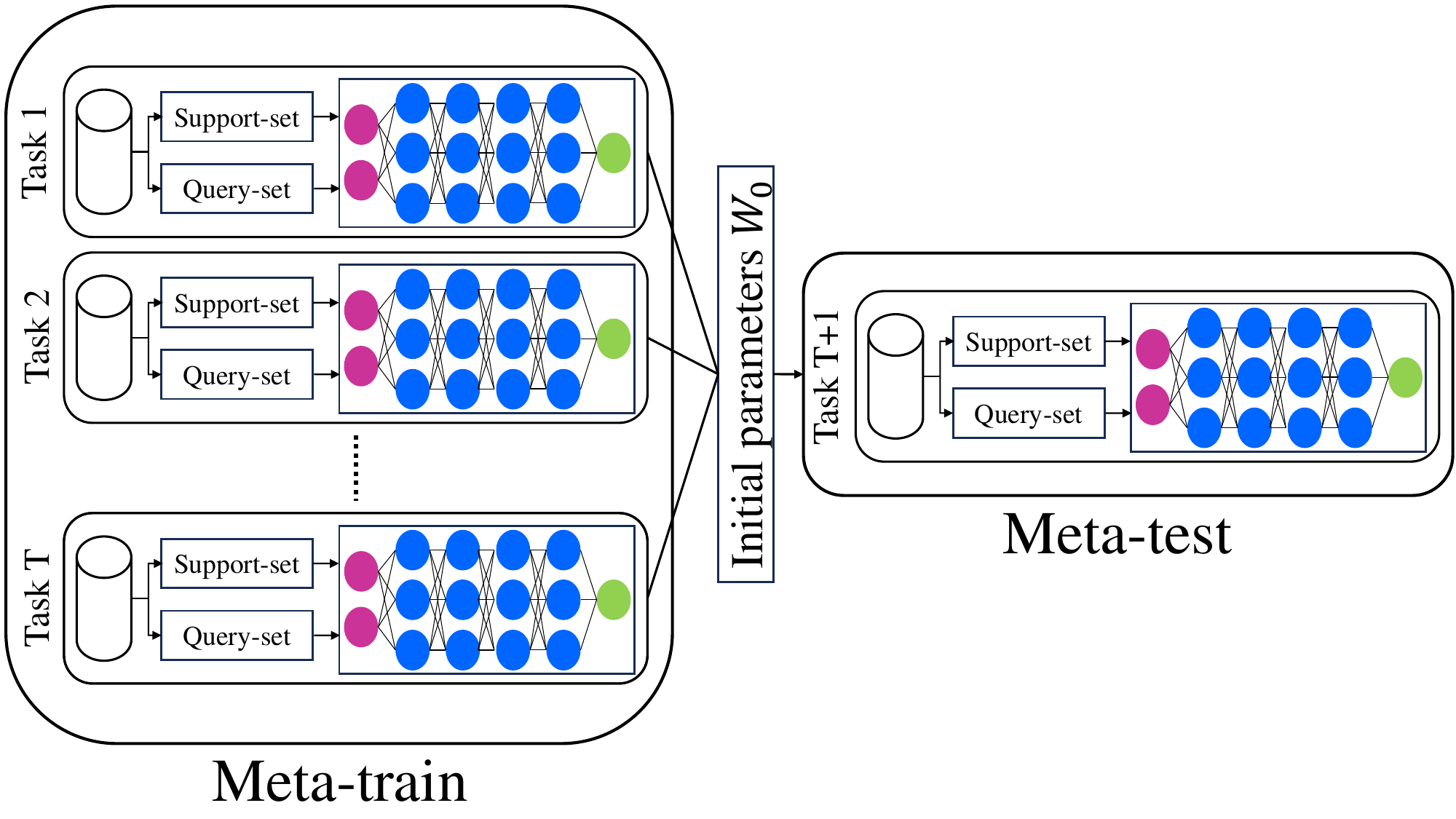} \vspace{2mm}
    \caption{Meta-learning architecture. It consists of i) a meta-train that has a support set, which samples meta-tasks for training, and ii) a query set that tests the meta-learning model on new, unseen tasks.} 
    \label{Meta_Learning}
\end{figure}

\emph{Meta-learning}~\cite{finn2017modelagnostic}, known as \emph{learning-to-learn}, utilizes learning across multiple tasks to infer an optimized learning policy for a new task. As illustrated in Fig.~\ref{Meta_Learning}, \emph{Model-agnostic meta-learning (MAML)}~\cite{finn2017modelagnostic} is a well-known meta-learning algorithm that aims at achieving faster convergence through a few SGD steps by optimizing a common initial parameter $W_0$ across multiple tasks. Assume a task distribution $p(\tau)$ and $T$ tasks ${\tau_1, \cdots, \tau_T}$ sampled randomly and in an i.i.d. fashion from the distribution. These tasks are fed to the \emph{meta-training} phase, where each task $\tau_i$ comprises a \emph{support-set}, which is used for training, and a \emph{query-test}, which is utilized for testing.

\begin{figure*}[t!]
    \centering
    \includegraphics[width=1.5\columnwidth,trim={0 0 0 0},clip]{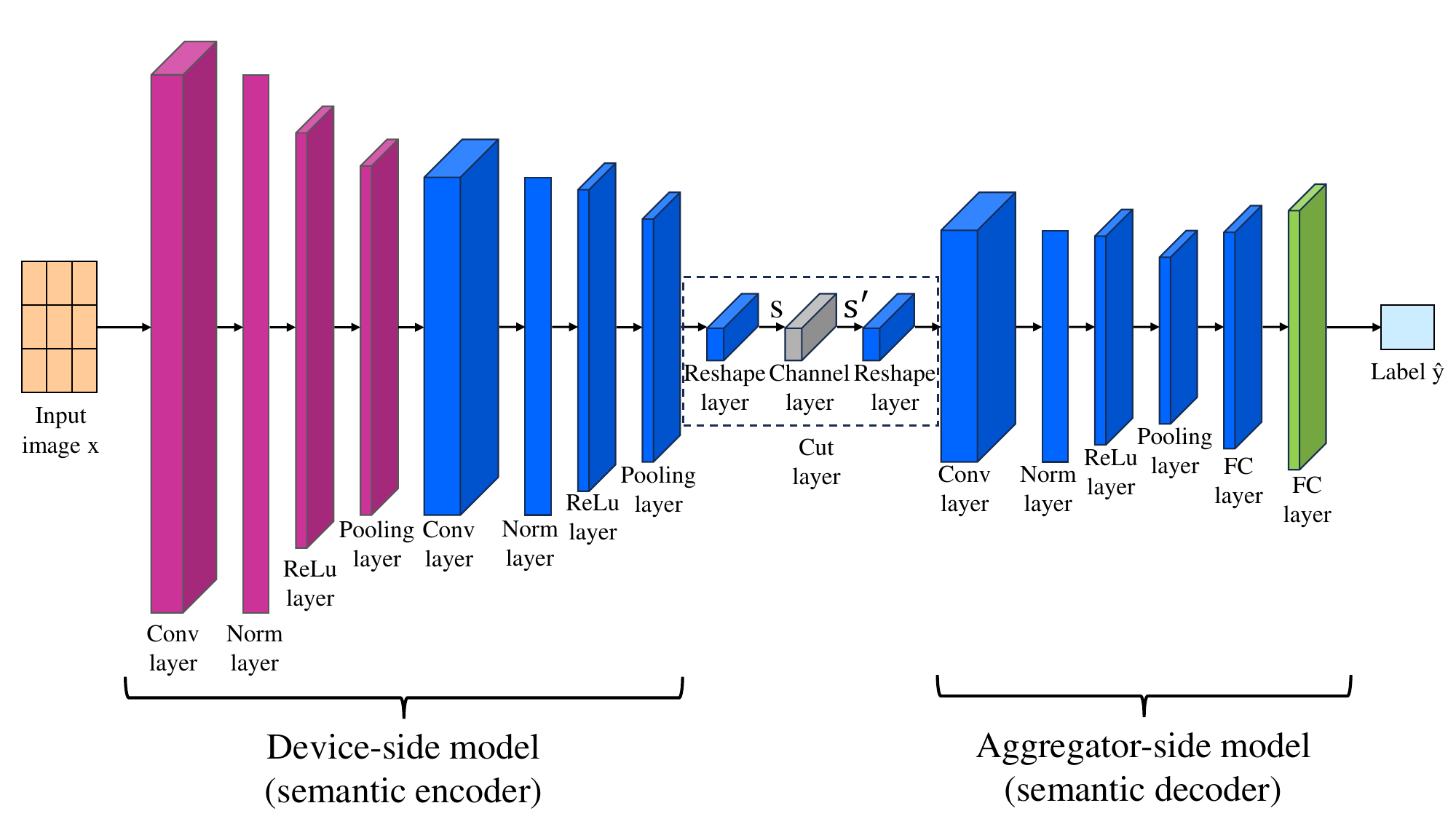} \vspace{2mm}
    \caption{The proposed Semantic-MSL architecture. The model consists of multiple convolutional layers, each followed by a normalization layer, a ReLu layer, and a pooling layer. At the end, multiple fully connected layers output the predicted class. At the cut layer, a reshape layer corrects the size of the smashed data to be transmitted through the channel.} 
    \label{CNN_Model}
\end{figure*}

Consider a model with parameter $\mathbf{W}$ and a loss function $\mathcal{L}_{\mathbf{W}} (\tau_i)$ adopted to train each task $\tau_i$
\begin{equation}
    \label{task_update}
    \mathbf{W} \leftarrow \mathbf{W} - \eta \nabla_{\mathbf{W}} \mathcal{L}_{\mathbf{W}} (\tau_i).
\end{equation}
The same update is applied to all the sampled tasks, and the meta-loss is calculated using the updated parameters for each task
\begin{equation}
    \label{meta_loss}
    \mathcal{L}_{\text{meta}} = \sum_{i=1}^{T} \mathcal{L}_{\mathbf{W}} (\tau_i).
\end{equation}
The meta-optimization is performed collaboratively across all the tasks by using SGD with the meta-loss
\begin{equation}
    \label{meta_optimization}
    \mathbf{W} \leftarrow \mathbf{W} - \beta \nabla_{\mathbf{W}} \mathcal{L}_{\text{meta}},
\end{equation}
where $\beta$ is the meta-learning rate.

 The \emph{meta-testing} phase adopts a new task $\tau_{T+1}$ composed of a support set and a query test. The optimized parameters $\mathbf{W}$ are the initial parameters for the new sampled task, and throughout a few SGD steps on the optimized parameters, the model converges. The number of samples in the support set (in both meta-training and meta-testing) is called \emph{shots $K$}, where using a small number of shots is referred to as \emph{few-shot learning}.

\section{Semantic Meta-Split Learning Framework}\label{sec:proposed}
This section presents the novel Semantic-MSL learning-based framework for wireless image transmission. This approach combines split learning with meta-learning to formulate a low computational device-side architecture wireless image classification using a small number of images.
As shown in Fig.~\ref{System_Model}, the device feeds the semantic information of an image forward through to the device-side network, i.e., the semantic encoder. Then, the receiver estimates the image class through the aggregator-side network, i.e., the semantic decoder. In addition, we assume a few-shot classification problem, where only a small number of images are available for each task.

To this end, for a particular task $\tau_i$, the semantic decoder with parameters $\mathbf{W^S}$ and a loss function $\mathcal{L}_{\mathbf{W^C},\mathbf{W^S}} (\mathbf{y},\hat{\mathbf{y}};\tau_i)$ is trained at the aggregator-side as
\begin{equation}
    \label{task_update_server}
    \mathbf{W^S} \leftarrow \mathbf{W^S} - \eta \nabla_{\mathbf{W^S}} \mathcal{L}_{\mathbf{W^C},\mathbf{W^S}} (\mathbf{y},\hat{\mathbf{y}};\tau_i).
\end{equation}
Similarly, the semantic encoder with parameter $\mathbf{W^C}$ would be trained at the device-side as 
\begin{equation} 
    \label{task_update_client}
    \mathbf{W^C} \leftarrow \mathbf{W^C} - \eta \nabla_{\mathbf{W^C}} \mathcal{L}_{\mathbf{W^C},\mathbf{W^S}} (\mathbf{y},\hat{\mathbf{y}};\tau_i),
\end{equation}
where the gradient $\nabla_{\mathbf{W^C}} \mathcal{L}_{\mathbf{W^C},\mathbf{W^S}} (\mathbf{y},\hat{\mathbf{y}};\tau_i)$ is transmitted from the aggregator to the device through the wireless channel. Afterward, the newly updated parameters $\mathbf{W^C}$ and $\mathbf{W^S}$ are used to estimate the meta-split loss across the tasks used for meta-split training
\begin{equation}
    \label{meta_loss_split}
    \mathcal{L}_{\text{meta-split}} = \sum_{i=1}^{T} \mathcal{L}_{\mathbf{W^C},\mathbf{W^S}} (\mathbf{y},\hat{\mathbf{y}};\tau_i).
\end{equation}
 Finally, the initial parameters are updated using the meta-split update
\begin{align}
    \label{meta_split_optimization_Client}
    &\mathbf{W^C} \leftarrow \mathbf{W^C} - \beta \nabla_{\mathbf{W^C}} \mathcal{L}_{\text{meta-split}},\\ \label{meta_split_optimization_Server}
    &\mathbf{W^S} \leftarrow \mathbf{W^S} - \beta \nabla_{\mathbf{W^S}} \mathcal{L}_{\text{meta-split}}.
\end{align}
The aforementioned steps are repeated till convergence. During testing, a few SDG steps are carried out on the semantic encoder and semantic decoder with a few shots ($K$) on a new task $\tau_{T+1}$.

Fig.~\ref{CNN_Model} illustrates the proposed \emph{convolutional neural network (CNN)} for both semantic encoder and semantic decoder. The model consists of multiple convolutional layers, each followed by a normalization layer, a ReLu layer, and a pooling (max) layer. After the convolutional layers, multiple fully-connected (FC) layers are followed by a softmax layer that outputs the estimated label (class). The cut layer controls the semantic encoder and the semantic decoder. The convolution layers are feature extraction layers, and thus, the later the cut layer, the lower the dimension of the transmitted semantic message $\mathbf{s}$. In contrast, the earlier the cut layer, the lower the computations performed at the device side. This illustrates the trade-off between the rate of encoding and the computational complexity. \textbf{Algorithm~\ref{Meta_Split_Alg}} summarizes the proposed Semantic-MSL framework.

\begin{algorithm}[!t]
\SetAlgoLined

\textbf{Input:} Semantic learning rate $\eta$, meta-learning rate $\beta$, number of tasks $T$, and number of training epochs $E$

\textbf{Output:} Initial parameters $\mathbf{W^C}$ and $\mathbf{W^S}$

Initialize network parameters $\mathbf{W^C}$ and $\mathbf{W^S}$

\For{\text{epochs} $e$ in $\{1, \cdots, E\}$}{

Sample $T$ tasks from the distribution $p(\tau)$

\For{\text{each sampled task} in $\{\tau_1, \cdots, \tau_T\}$}{

Sample $K$ shots for each class in each task

Update the model parameters $\mathbf{W^C}$ and $\mathbf{W^S}$ using~\eqref{task_update_server} and~\eqref{task_update_client}
}
    
Calculate the meta-split loss $\mathcal{L}_{\text{meta-split}}$ using~\eqref{meta_loss_split}

Update the initial parameters $\mathbf{W^C}$ and $\mathbf{W^S}$ using~\eqref{meta_split_optimization_Server} and~\eqref{meta_split_optimization_Client}

}

\textbf{Return} model initial parameters $\mathbf{W^C}$ and $\mathbf{W^S}$

\caption{The proposed Semantic-MSL algorithm.}
\label{Meta_Split_Alg}  
\end{algorithm}

\section{Key Performance Metrics}\label{Sec:KPI}

This section presents the key performance metrics leveraged to evaluate the proposed model compared to other baseline schemes. For instance, classification accuracy is defined as the total number of true classified images divided by the total number of images. The precision, recall, and f1-score are considered the fundamental metric in evaluating classification problems~\cite{9504554}. However, these metrics lack measuring the degree of uncertainty and the model's trustworthiness, which is usually common in deep learning models with limited access to data points~\cite{bates2021distribution}.

\emph{Conformal prediction (CP)}~\cite{angelopoulos2022gentle} is a framework that quantifies the degree of uncertainty in predictions of predictive models (such as neural networks). CP calibrates the models by generating prediction sets instead of a single prediction. Consider an input $\mathbf{x}$ corresponding to an output $\mathbf{y}$ from the set of outputs $\mathcal{Y}$. Relying on the fact that the training and testing of input data are exchangeable, CP produces a subset of the output set, i.e., \emph{prediction set $\Gamma$}, which contains the true output with a probability $1-\alpha$, where $\alpha \in [0,1 ]$~\cite{angelopoulos2022gentle}. 

CP's prediction set is evaluated in terms of \emph{coverage} $\texttt{cov}(\Gamma)$ and \emph{inefficiency} $\texttt{ineff}(\Gamma)$. The former measures the probability of the true label in the prediction set, whereas the latter is the size of the prediction set. The coverage and the inefficiency are formulated as
\begin{align}
    \label{Cov_eq}
    \texttt{cov}(\Gamma) &= P(y \in \Gamma), \\ \label{ineff_eq}
    \texttt{ineff}(\Gamma) &= \mathbb{E} \: \big[ | \Gamma | \big],
\end{align} 
where the average $\mathbb{E} \: [ \cdot ]$ is taken over the given data points. A model that generates a prediction set that is equivalent to the entire output set yields the maximum possible coverage $ \texttt{cov}(\Gamma) = 1$ at the cost of having the worst inefficiency $\texttt{ineff}(\Gamma) = |\mathcal{Y}|$. In contrast, a model that generates an empty prediction set achieves the best possible inefficiency $\texttt{ineff}(\Gamma) = 0$ at the cost of the coverage $ \texttt{cov}(\Gamma) = 0$. Hence, a well-calibrated model would maintain the trade-off between coverage and inefficiency by achieving no lower than $1-\alpha$ coverage with a relatively low inefficiency.

\emph{Validation-based CP (VB-CP)} is a class of CP in which the data points are divided into calibration data set $D^{c}$ and validation data $D^{v}$. Using the calibration data, we calculate the \emph{nonconformity} (NC) score, which is a function that describes how poor an output is for a given input. For instance, a well-known NC score is
\begin{equation}
    \label{NC_score_fn}
    \text{NC}(D^{c}) = 1 - \text{Softmax}(D^{c}).
\end{equation}
Then, we compute the $q^{\text{th}}$ quantile $\hat{q} = \frac{\ceil*{(n+1)(1-\alpha)}}{n}$ of the NC scores of the calibration data, where $n$ is the size of the calibration data. For the validation data, the prediction set is generated, such that it includes all the classes that have NC-scores smaller than or equal to the empirical quantile~\cite{gibbs2021adaptive}
\begin{equation}
    \label{pred_set}
    \Gamma = \{ \mathbf{y}^{\prime} \in \mathcal{Y} \: | \: \text{NC}(\mathbf{x},\mathbf{y}^{\prime};D^{v}) \leq  q^{\text{th}} \: \text{NC}(D^{c}) \},
\end{equation}

\section{Numerical Results}\label{sec:results}

\begin{table}[t!]
\centering
\caption{Simulation parameters and hyperparameters.}
\label{Sim_Parameters}
\begin{tabular}{cc|cc}
\toprule
\textbf{Parameter}                                    & \textbf{Value} & \textbf{Parameter}                                    & \textbf{Value} \\ \midrule
\midrule
$T$ & $20$ & $Y$ & $10$ \\
$M$ & $20$ & $K$ & $5$ \\
$\alpha$ & $0.1$ & epochs $E$ & $1000$ \\
Conv. layers & $3$ layers ($64$ neurons) & FC. layers & $2$ layers ($64$ neurons) \\
Kernel size & $3$ & stride & $2$ \\
$\eta$ & $0.001$ & $\beta$ & $0.01$ \\
Optimizer & Adam & loss & CrossEntropy \\

\bottomrule
\end{tabular} 
\end{table}

\begin{figure}[t!]
    \centering
    \includegraphics[width=1\columnwidth,trim={0 0 0 0},clip]{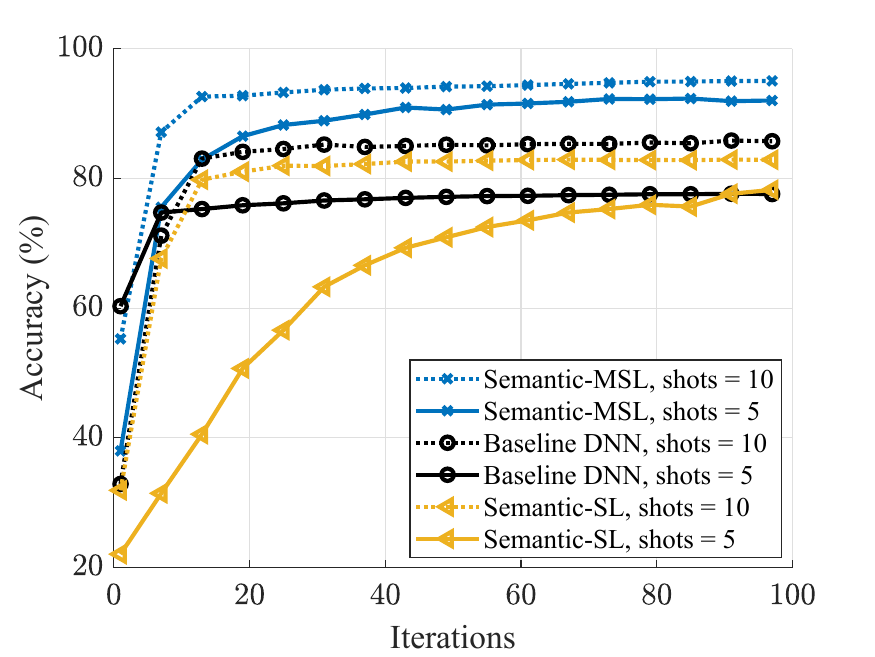} \vspace{2mm}
    \caption{The convergence of the accuracy as a function of the SGD steps (iterations) of the proposed Semantic-MSL compared to the Semantic-SL scheme.} 
    \vspace{0mm}
    \label{Iterations}
\end{figure}

This section presents the simulation results of the proposed Semantic-MSL framework. We use $3$ convolution neural network with $64$ neurons each followed by two fully connected neural networks with two hidden layers of size $64$ and ReLU activation functions. The experiments are implemented using Pytorch on a single NVIDIA Tesla V100 GPU. We implement the experiments using Omniglot data set~\cite{lake2015human}, which consists of $1623$ classes of letters from $50$ different languages. Each class contains $20$ different hand-written images. Table~\ref{Sim_Parameters} illustrates the parameters used in the simulation. 
The proposed Semantic-MSL model is compared to Semantic-split learning (Semantic-SL), which only uses split learning, and a baseline DNN model, where the whole model is built and trained at the device. Moreover, we test the proposed model while adjusting the number of training tasks, number of shots, and position of the cut layer.

\begin{figure}[t!]
    \centering
    \subfloat[Accuracy for different number of shots.]{\includegraphics[width=0.48\textwidth,trim={0 0 0 0},clip]{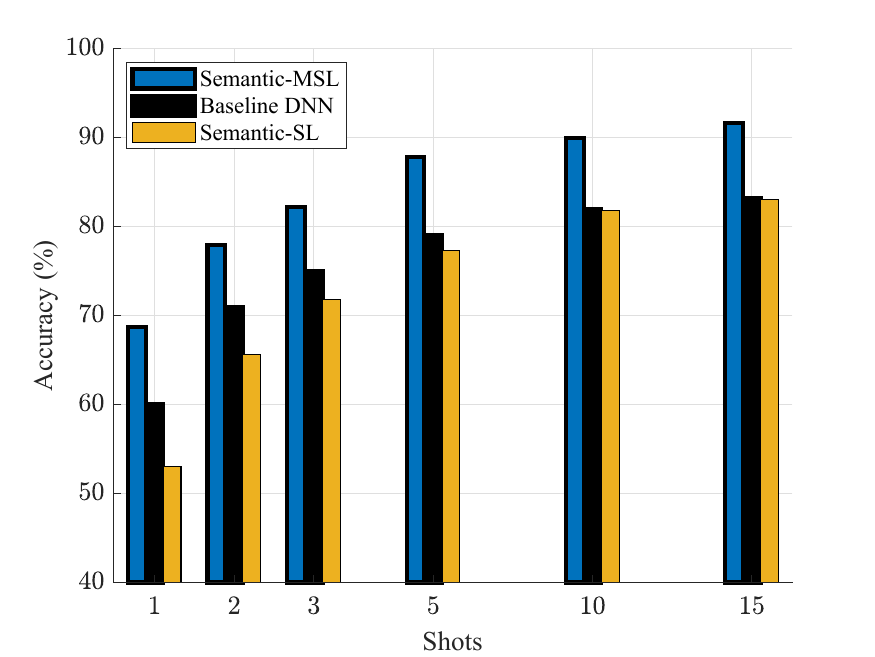} \label{Shots}} 
    \hskip -2.28ex
    \subfloat[Accuracy for different number of tasks]{\includegraphics[width=0.48\textwidth,trim={0 0 0 0},clip]{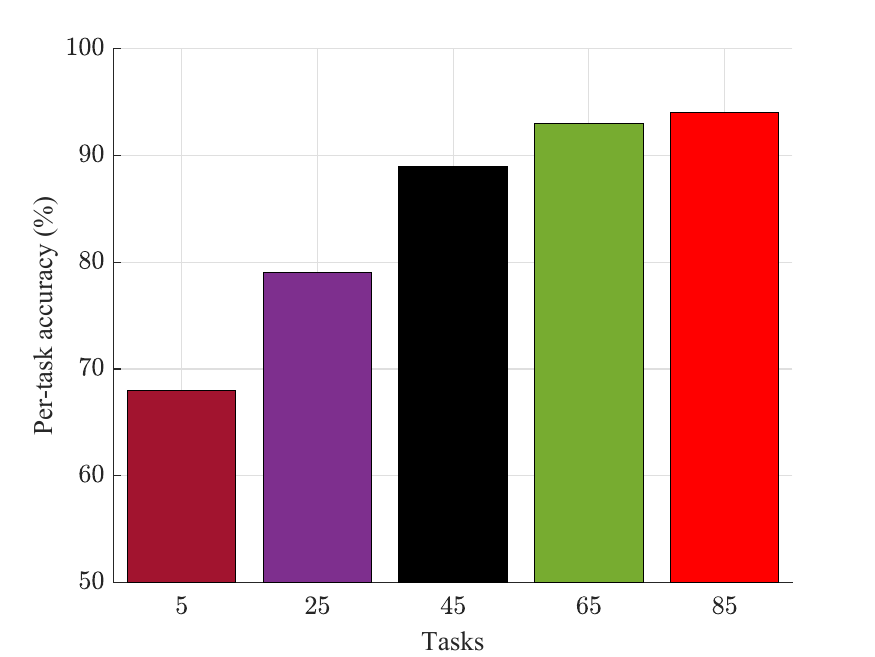} \label{Tasks}} 
    \caption{The classification accuracy is a function of both the number of shots (number of images used in training) and the number of meta-tasks. In (a), compares the accuracy of Semantic-MSL, Semantic-SL, DNN schemes as function of the number of shots. In (b), the Semantic-MSL scheme's per-task accuracy is plotted as a function of the number of meta-tasks.}
    \label{Shots_Tasks}
\end{figure}

In Fig.~\ref{Iterations}, we report the achievable classification accuracy after a given number of iterations of SGD steps using $5$-shots of images from each class. In this experiment, Semantic-MSL is trained on $65$ different tasks and compared to the baseline DNN and Semantic-SL schemes. We can notice that the Semantic-MSL converges to an optimum accuracy of $95 \%$ after only $20$ SGD steps outperforming both DNN and SL. This occurs due to the enhanced initialization of the MAML algorithm obtained from meta-training of $65$ tasks. In contrast, the Semantic-SL scheme reaches less than $60 \%$ accuracy after $20$ SGD steps, which is a bit lower the DNN, despite the superiority of SL over DNN in terms of reduced power consumption.  In addition, it spends $100$ SGD steps to only reach a sub-optimum accuracy of $82 \%$. This highlights the benefits of meta-learning in utilizing experience from other tasks to improve and speed up learning new tasks.

\begin{figure*}[t!]
    \centering
    \subfloat[Coverage]{\includegraphics[width=0.48\textwidth,trim={0 0 0 0},clip]{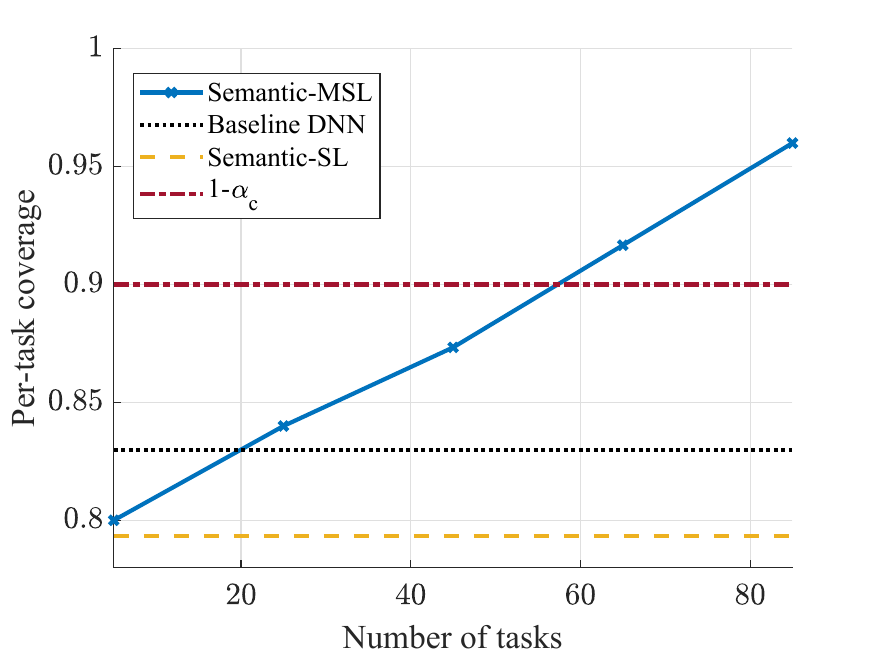} \label{Coverage}} 
    \hskip -2.28ex
    \subfloat[Inefficiency]{\includegraphics[width=0.48\textwidth,trim={0 0 0 0},clip]{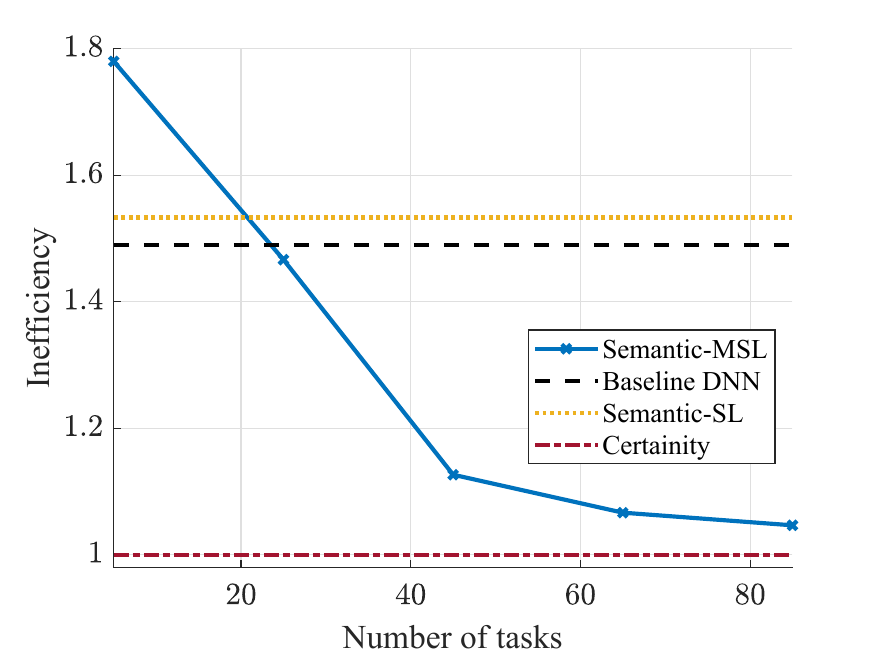} \label{Inefficiency}} 
    \caption{Coverage and inefficiency of the Semantic-MSL and Semantic-SL schemes across different tasks.}
    \label{Cover_Ineffic}
\end{figure*}

Fig.~\ref{Shots_Tasks} depicts the effect of the number of image shots used in training for each class in a particular task on the classification accuracy for both Semantic-MSL and Semantic-SL schemes. In addition, we show the effect of the number of meta-training tasks on the classification accuracy for the proposed Semantic-MSL scheme. Herein, the number of training iterations is fixed to $30$ SGD steps. In Fig.~\ref{Shots}, we observe that increasing the number of training shots improves the classification accuracy. Interestingly, the proposed Semantic-MSL scheme achieves more than $91 \%$ accuracy with only $5$ shots (we keep the number of meta-tasks fixed to $45$ tasks). In contrast, Semantic-SL and DNN fail to exceed $83 \%$ accuracy even with 15 shots. Moreover, Fig.~\ref{Tasks} illustrates that increasing the number of meta-training tasks enhances accuracy. We can notice that classification accuracy does not improve much when the number of tasks exceeds $65$. Increasing the number of tasks over $85$ does not provide any noticeable gain in the classification accuracy as saturation behavior is noticed.

The next experiment investigates the uncertainty of the proposed Semantic-MSL model compared to Semantic-SL and DNN using the discussed conformal prediction metrics as depicted in Fig.~\ref{Cover_Ineffic}. The first aspect of conformal prediction metrics is the coverage, which measures the probability of the true class in the prediction set $\Gamma$, shown in Fig.~\ref{Coverage}. Setting $\alpha = 0.1$, we observe that increasing the number of tasks enhances Semantic-MSL coverage as it approaches $1$ for a larger number of meta-tasks. In contrast, the coverage of Semantic-SL and DNN are always constant and inferior to Semantic-MSL. To obtain coverage larger than or equal to $1-\alpha$, i.e., $0.9$, meta-learning requires $55$ tasks or more. In Fig.~\ref{Inefficiency}, we observe the inefficiency, which is the average size of the prediction set $\Gamma$. Semantic-SL has a constant set size of $1.53$ compared to Semantic-MSL, which is affected by the number of tasks. The higher the number of meta-tasks, the closer the model to predict certainly (i.e., approaches inefficiency of $1$).

\begin{table}[t!]
\centering
 \caption{Semantic-MSL training duration, energy consumption, CO2 emissions, and the coverage accuracy for different meta-tasks.}
\label{tab:Tasks}
\begin{tabular}{@{}ccccc@{}}
\toprule
\textbf{Tasks} & \textbf{Duration (s)} & \textbf{Energy (Wh)} & \textbf{CO2 emissions (g)}  & \textbf{Coverage}\\ \midrule
\midrule
$\mathbf{5}$ & $336.72$ & $7.77$ & $1.10$ & $0.800$ \\
\hline 
$\mathbf{25}$ & $1659.12$ & $35.48$ & $5.84$ & $0.840$ \\
\hline 
$\mathbf{45}$ & $2892.28$ & $72.28$ & $10.25$ & $0.873$ \\
\hline 
$\mathbf{65}$ & $3819.81$ & $96.45$ & $13.68$ & $0.917$ \\
\hline 
$\mathbf{85}$ & $6041.63$ & $217.97$ & $30.92$ & $0.960$ \\
\bottomrule
\end{tabular}
\end{table}

To elaborate further on the computational complexity of the proposed model, we utilize ECO2AI\footnote{ECO2AI~\cite{budennyy2022eco2ai} is an open-source python library that tackles the CPU and GPU consumption and estimates the equivalent CO2 emissions. It aims to achieve models with low computational costs.} tool to estimate the computational costs using different training tasks and different split layer configurations. Table~\ref{tab:Tasks} reports the time complexity, computational energy consumption, and the accuracy coverage of the proposed Semantic-MSL scheme while adjusting the number of tasks in the MAML algorithm. We notice that the training duration, energy consumption, and CO2 footprint increase as meta-tasks increase. Similarly, the coverage accuracy is enhanced by increasing the number of meta-tasks. To this end, setting the number of tasks to $[45 - 65]$ renders relatively high accuracy and coverage while being conservative over the training duration and the required computational energy.

\begin{table*}[t!]
\centering
    \caption{Reporting the computation and communication energy consumption during training, 
            as well as the size of the transmitted data and device inference time during testing the Semantic-MSL 
            model while adjusting the position of the cut layer compared to the baseline model.}
\label{tab:COMP}
\begin{tabular}{@{}ccccc@{}}
\toprule
\textbf{Model / Cut-layer} & $\:$ \textbf{Computation energy (Wh)} $\:$ & $\:$ \textbf{Communication energy (Wh)} $\:$ & $\:$ \textbf{Size (KB)} $\:$ & $\:$ \textbf{Inference time (ms)}\\ \midrule
\midrule \\

\textbf{Baseline} & $44.66$ & $-$ & $-$ & $889.54$ \\
\\ \hline \\
\textbf{$\mathbf{3}$-Conv} & $41.20$ & $3.45 \times 10^{-3}$ & $0.248$ & $770.54$ \\
\\ \hline \\
\textbf{$\mathbf{2}$-Conv} & $35.48$ & $4.51 \times 10^{-2}$ & $6.200$ & $479.25$ \\
\\ \hline \\
\textbf{$\mathbf{1}$-Conv} & $29.33$ & $7.83 \times 10^{-1}$ & $14.912$ & $287.64$ \\
\\ \bottomrule
\end{tabular}
\end{table*}

Finally, in Table~\ref{tab:COMP}, we report the computation energy, the size of the transmitted semantic message, the communication energy, and the deployment time (inference time) of the device-side model while adjusting the position of the cut-layer. The basleine DNN has the highest computation energy and inference time. Setting the cut layer after the third convolutional layer ($3$-Conv) leaves the aggregator-side model with only the fully connected layers. In that case, the size of the transmitted message is small ($0.248$ KB) since the convolutional layers are feature extractors, and the communication energy is relatively small compared to other cut-layer positions. However, it loads the device-side model with higher computational energy and a larger deployment time. In contrast, shifting the cut-layer position to be after the second convolutional layer ($2$-Conv) or the first convolutional layer ($1$-Conv) reduces the required computational energy and deployment time at the devices at the costs of increasing the size of the transmitted message and the communication energy. This highlights the trade-off between computational energy, deployment time, communication energy, and overhead. Note that the computation energy is dominant over the communication energy in all cases; therefore, reducing the computation energy by moving the cut layer closer to the device size (i.e, 1-Conv) reduces the overall energy consumption of the device. However, this comes at the cost of higher communication overhead and the risk of privacy incursions, which is an open area for future research. 

\section{Conclusions}\label{sec:conclusions} 
In this paper, we proposed Semantic-MSL as a novel Tiny-ML framework for few-shot wireless image classification. The devices transmit the semantics of an image over a wireless channel to the aggregator, which predicts the true class of the image via meta-training over several classification tasks. Numerical results depicted the benefits of combining meta-learning with split-learning, which achieves $20 \%$ higher classification accuracy than using only split-learning, while using fewer training shots simultaneously. Regarding CP aspects, Semantic-MSL renders high coverage and low inefficiency. Moreover, the overall energy consumption of the proposed Semantic-MSL is significantly reduced by moving the cut layer closer to the device side. 

In summary, Semantic-MSL outperforms baseline DNN and Semantic-SL in termas of accuracy, CP accuracy, CP inffiency, and energy consumption. The proposed framework can be further adopted in tasks other than wireless image classification, such as image reconstruction and reinforcement learning. Exploiting meta-conformal prediction in optimizing the split learning model is left for future research.

\appendices 
%


\bibliographystyle{IEEEtran}
\bibliography{IEEEabrv,references}
\end{document}